\documentclass[runningheads]{llncs}
\usepackage[T1]{fontenc}
\usepackage{graphicx, verbatim}

\usepackage{amsmath}   
\usepackage{amssymb}   
\usepackage{bm}        

\usepackage{multirow}
\usepackage{booktabs}
\usepackage{array}
\usepackage{siunitx}
\usepackage{caption}
\usepackage{diagbox}
\usepackage{adjustbox}

\usepackage{pifont}
\usepackage{xcolor}
\usepackage[ruled,vlined]{algorithm2e} 
\usepackage{makecell}  

\def\ie{\textit{i.e.}}
\def\eg{\textit{e.g.}}

\usepackage[colorlinks,linkcolor=red, anchorcolor=blue, citecolor=teal, urlcolor=magenta]{hyperref}

\newcommand{\myparagraph}[1]{\vspace{1pt}\noindent{\bf{#1}}~~}

\begin{document}
\title{OTSurv: A Novel Multiple Instance Learning Framework for Survival Prediction with Heterogeneity-aware Optimal Transport}

\author{
Qin Ren$^{1}$\quad
Yifan Wang$^{1}$\quad
Ruogu Fang$^{3}$\quad
Haibin Ling$^{1}$\quad
Chenyu You$^{1,2}$\quad\\
}

\authorrunning{Q. Ren et al.}
\institute{$^1$Department of Computer Science, Stony Brook University \\
$^2$ Department of Applied Mathematics \& Statistics, Stony Brook University \\
$^3$ Department of Biomedical Engineering, University of Florida \\
\email{\{qinren,cyou\}@cs.stonybrook.edu}
}

\maketitle              

\begin{abstract}

Survival prediction using whole slide images (WSIs) can be formulated as a multiple instance learning (MIL) problem. However, existing MIL methods often fail to explicitly capture pathological heterogeneity within WSIs, both globally -- through long-tailed morphological distributions, and locally -- through tile-level prediction uncertainty. 
Optimal transport (OT) provides a principled way of modeling such heterogeneity by incorporating marginal distribution constraints. Building on this insight, we propose \textbf{OTSurv}, a novel MIL framework from an optimal transport perspective. Specifically, OTSurv formulates survival predictions as a heterogeneity-aware OT problem with two constraints: (1) \textit{global long-tail constraint} that models prior morphological distributions to avert both mode collapse and excessive uniformity by regulating transport mass allocation, and (2) \textit{local uncertainty-aware constraint} that prioritizes high-confidence patches while suppressing noise by progressively raising the total transport mass.  We then recast the initial OT problem, augmented by these constraints, into an unbalanced OT formulation that can be solved with an efficient, hardware-friendly matrix scaling algorithm. Empirically, OTSurv sets new state-of-the-art results across six popular benchmarks, achieving an absolute 3.6\% improvement in average C-index. In addition, OTSurv achieves statistical significance in log-rank tests and offers high interpretability, making it a powerful tool for survival prediction in digital pathology. Our codes are available at \url{https://github.com/Y-Research-SBU/OTSurv}.

\end{abstract}


\section{Introduction}

Survival prediction, which estimates patient-specific time-to-event probabilities (\ie, overall survival), is a critical oncology task essential for optimizing therapeutic strategies \cite{song2024analysis}. In clinical practice, it mainly relies on pathologists’ meticulous analysis of tissue slides. However, variations in tissue composition and structure across different regions pose significant challenges in identifying prognostic patterns \cite{campanella2019clinical,gerlinger2012intratumor,mcgranahan2017clonal,you2023rethinking,you2024mine,ma2025pathbench,jin2024hmil}. In particular, pathologists need to detect and interpret small or ambiguous regions that are crucial for survival outcomes. Capturing this heterogeneity is inherently challenging, and even slight missteps can lead to incomplete prognostic assessments that ultimately compromise patient survival.

In digital pathology, neural networks have shown great promise in WSI-based survival prediction, but their ultra-high resolution (\eg,~$10^5 \times 10^5$ pixels) incurs high computational and annotation costs, rendering fully supervised learning impractical \cite{xu2024whole}. Thus, weakly supervised methods, particularly Multiple Instance Learning (MIL), have become the \textit{de facto} solution, which treats each WSI as a bag of instances with only a bag-level survival label \cite{xiang2023exploring,zhang2022dtfd,li2023task,shao2021transmil}. Analogous to how pathologists analyze tissue heterogeneity, MIL methods are required to effectively capture this variability to ensure that crucial survival-related information is not lost. Specifically, survival-related heterogeneity manifests at two scales. At \textit{global} level, it involves capturing sparse patterns within the long-tailed distribution of patch types \cite{hou2016patch}. At \textit{local} level, it focuses on eliminating the predictive uncertainty of each patch \cite{ren2023iib}. However, this dual-scale heterogeneity remains under-explored in current MIL approach: some completely ignore heterogeneity (\eg, Mean/Max Pooling), some focus only on single level (\textit{global} \cite{songmultimodal,lin2023interventional} or \textit{local} \cite{tang2023multiple,ilse2018attention,li2023task,shao2021transmil,xiang2023exploring}), while others attempt to tackle both but without providing explicit guidance \cite{yao2020whole}. Additionally, computational limits force MIL models to sample tiles randomly, risking the omission of critical prognostic regions and hindering heterogeneity modeling.

In this work, we aim to address the following question: \textit{how can MIL-based models explicitly account for both global and local pathological heterogeneity?}  We note that this heterogeneity in MIL essentially manifests as feature distribution variability, with global and local heterogeneity reflected in prototype-level and instance-level feature distributions, respectively. Optimal Transport (OT) \cite{villani2009optimal,khamis2024scalable}, as a mathematically interpretable framework that optimally aligns two distributions while minimizing transport cost, naturally fits this setting. By incorporating two marginal distribution constraints, OT provides a principled mechanism to explicitly model dual-scale heterogeneity.

Building on this insight, we propose \textbf{OTSurv}, an \textbf{\underline{OT}}-based MIL framework for \textbf{\underline{surv}}ival prediction from an optimal transport perspective. OTSurv models survival prediction as an OT problem by aggregating a variable-sized set of instance features and mapping them onto a fixed-size set of trainable survival tokens, which is used for reference. This process explicitly accounts for both global and local heterogeneity through two tailored marginal distribution constraints. For \textit{global heterogeneity} (long-tailed distribution), we introduce \textbf{\underline{G}}lobal Long-tail \textbf{\underline{C}}onstraint ($\mathbf{C_G}$), which models prior morphological distributions to prevent mode collapse and excessive uniformity by regularizing transport mass allocation. For \textit{local heterogeneity} (difficult-to-predict patches), we propose \textbf{\underline{L}}ocal Uncertainty-aware \textbf{\underline{C}}onstraint ($\mathbf{C_L}$), which prioritizes high-confidence patches to suppress noise by progressively raising total transport mass. We then recast the augmented OT problem into an unbalanced OT formulation that can be solved with an efficient, hardware-friendly matrix scaling algorithm, by incorporating a virtual survival token to absorb unselected mass.

Our key contributions are:  (i) We propose \textbf{OTSurv}, a novel OT-based MIL framework for WSI survival prediction.  (ii) We design two tailored marginal constraints in OT to model global and local heterogeneity in WSIs. (iii) Experiments on 6 popular benchmarks show that OTSurv outperforms previous SOTA methods in C-index, achieves statistical significance in log-rank tests, and provides strong interpretability.

\begin{figure*}[t]
\begin{center}
\includegraphics[width=0.90\linewidth]{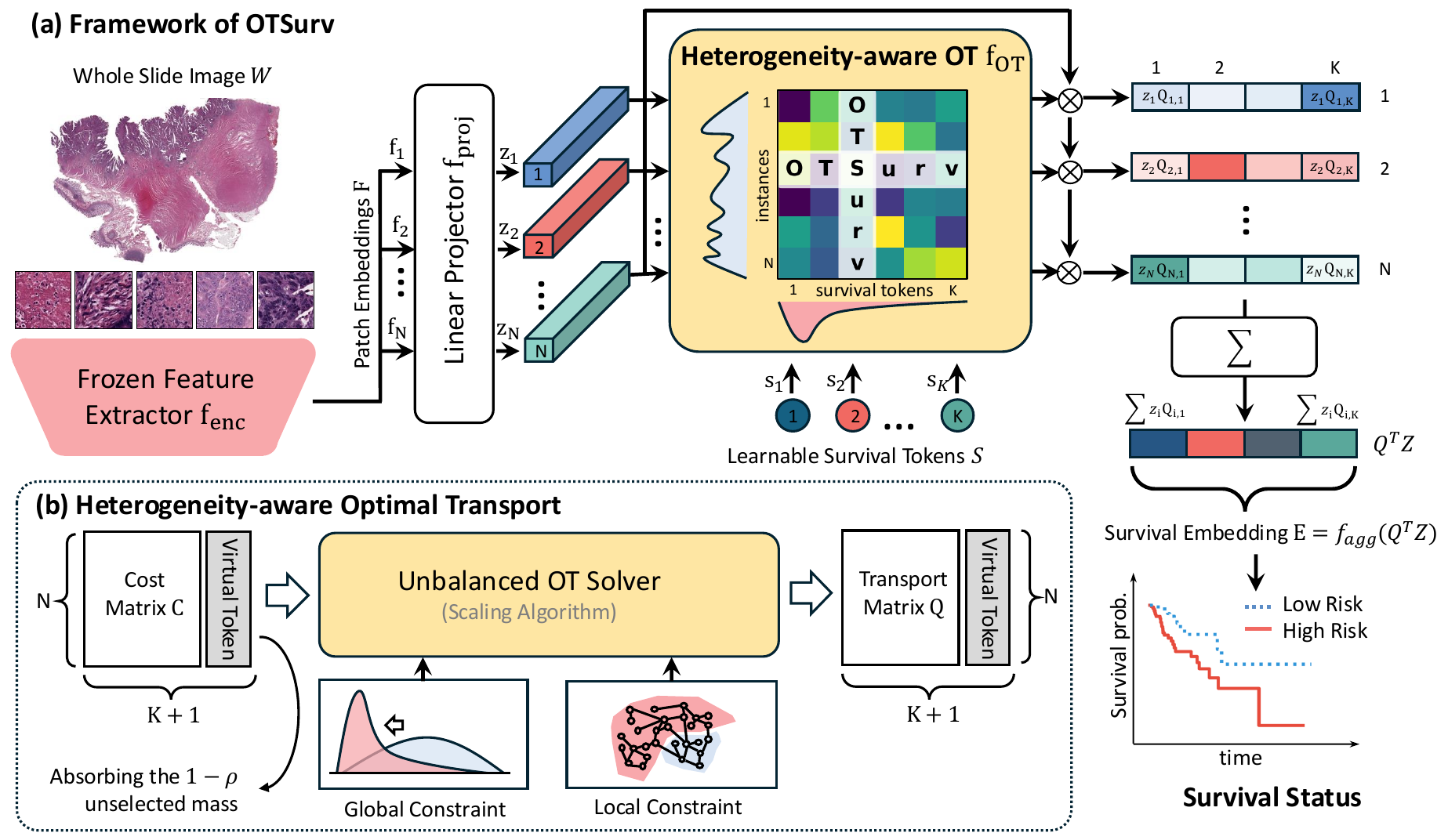}
\end{center}
\vspace{-15pt}
\caption{\textbf{Overview of the OTSurv framework.} (a) Framework of OTSurv: A WSI \(W\) is processed into patch features \(Z\), aligned with survival tokens \(S\) via heterogeneity-aware OT, and aggregated into a survival embedding \(E\). (b) heterogeneity-aware Optimal Transport: By extending the cost matrix with an additional column of zeros (\ie, virtual token),  the unbalanced OT solver can be used for solving heterogeneity-aware OT with global and local constraints.}
\label{fig_OTSurv}
\vspace{-10pt}
\end{figure*}

\section{Method}

\subsection{Overview}
\label{sec:framework}

As illustrated in Fig.~\ref{fig_OTSurv}, {OTSurv} reframes WSI-based survival prediction as a heterogeneity-aware optimal transport problem. It is composed of four distinct, yet interlocking, modules:
{(i)} \textit{WSI Decomposition}: A large-scale WSI $W$ is partitioned into $N$ non-overlapping patches $\{x_i\}_{i=1}^N$, denoted as \(x = \{x_i\}_{i=1}^N \in \mathbb{R}^{N \times c \times h \times w}\), where \(c\) is the number of color channels, and \(h \times w\) is the patch resolution; 
{(ii)} \textit{Feature Embedding}: A frozen encoder  $f_{\text{enc}}$  extracts patch-level features $F = f_{\text{enc}}(x) \in \mathbb{R}^{N \times D}$. These are then projected into a lower-dimensional latent space via a learnable linear projection, yielding instance embeddings $Z =  f_{\text{proj}}(F) \in \mathbb{R}^{N \times d}$;
{(iii)} \textit{Feature Aggregation}: At the \textbf{core} of our approach, the heterogeneity-aware OT module computes an optimal transport plan $Q = f_{\text{OT}}(C) \in \mathbb{R}^{N \times K}$, where the cost matrix $C \in \mathbb{R}^{N \times K}$ is computed using normalized Euclidean distance between $Z \in \mathbb{R}^{N \times d}$ and learnable survival tokens $S \in \mathbb{R}^{K \times d}$. The aggregated slide-level embedding is then obtained as $E = f_{\text{agg}} (Q^{\top} Z) \in \mathbb{R}^{d}$, where $Q^{\top} Z \in \mathbb{R}^{K \times d}$ and $f_{\text{agg}} (\cdot)$ is linear layer; and {(iv)} \textit{Risk Prediction}: finally, the aggregated embedding $E$ is fed into a linear predictor $f_{\text{pred}}$ to produce the final risk score $r = f_{\text{pred}}(E)\in \mathbb{R}$. The entire model is optimized by minimizing a Cox proportional hazards loss \cite{cox1972regression}.

\subsection{Heterogeneity-aware OT Problem}
\label{sec:hot}

In this section, we first overview OT theory, and then delve into an in-depth discussion of our proposed local and global OT marginal distribution constraints, which are pivotal to our framework.

\myparagraph{General OT Formulation.} To formulate heterogeneity-aware OT, we start with the OT problem, which aims to transport one distribution to another with minimal cost. 
Given a source distribution $\mu \in \mathbb{R}^{N \times 1}$, a target distribution $\nu \in \mathbb{R}^{K \times 1}$, and a cost matrix $C \in \mathbb{R}^{N \times K}$, our objective is to determine a transport matrix $Q \in \mathbb{R}^{N \times K}$ that minimizes:
\begin{equation}
\label{eq:1}
    \min_{Q \in \mathbb{R}_+^{N \times K}} \langle Q, C \rangle_F + F_1\left(Q \mathbf{1}_K,\, \mu \right) + F_2\left(Q^\top \mathbf{1}_N,\, \nu \right)
\end{equation}
where $\langle \cdot, \cdot \rangle_F$ denotes the Frobenius inner product, and $Q_{ij}$ represents mass transported from the $\mu_i$ to $\nu_j$. $F_1$ and $F_2$ enforce constraints on the marginal distributions of $Q$. When these constraints are specified as equalities (\ie, \( Q \mathbf{1}_K = \mu \), \( Q^\top \mathbf{1}_N = \nu \)), the formulation reduces to Kantorovich’s classical OT problem \cite{kantorovich1942transfer}. Alternatively, when $F_1$ and $F_2$ impose inequality constraints (\eg, using the KL divergence), the problem turns into unbalanced OT \cite{liero2018optimal}.

\myparagraph{Global Long-tail Constraint (\(\mathbf{C_G}\)).}  
WSIs exhibit a long-tailed tissue morphology, where dominant patterns prevail while rare, prognostically critical features are scarce. Without any constraint, the transport mass may collapse onto a single survival token, whereas enforcing a strict equality constraint forces a uniform mass distribution across tokens -- neither case captures the true long-tailed nature of the data. To address this, we impose a KL divergence constraint on the survival token mass \( Q^\top\mathbf{1}_N \in \mathbb{R}_+^{K \times 1} \) to match a desired long-tailed prior. This global constraint (\(\mathbf{C_G}\)) preserves tissue diversity by ensuring that every infrequent, yet critical, patterns are adequately represented for accurate survival prediction. Specifically, we apply a KL divergence penalty on \( F_2(\cdot) \) while temporarily assuming a uniform distribution for \( F_1(\cdot) \), where \( Q\mathbf{1}_K \in \mathbb{R}_+^{N \times 1} \). This leads to the following semi-relaxed OT formulation:
\begin{equation}
\label{eq:3}
\begin{aligned}
    \min_{Q \in \Pi} \quad & \langle Q, C \rangle_F + \lambda\, \mathrm{KL}\left(Q^\top \mathbf{1}_N \,\|\, \tfrac{1}{K} \mathbf{1}_K\right) \\
    \text{subject to} \quad & \Pi = \left\{ Q \in \mathbb{R}_+^{N \times K} \;\middle|\; Q \mathbf{1}_K = \tfrac{1}{N} \mathbf{1}_N \right\}
\end{aligned}
\end{equation}
where \( \lambda \) is a weighting factor controlling the KL divergence regularization.

\myparagraph{Local Uncertainty-aware Constraint ($\mathbf{C_L}$).}  In Eq.~\eqref{eq:3}, the \( F_1(\cdot) \) constraint \( Q \mathbf{1}_K = \frac{1}{N} \mathbf{1}_N \) treats all instances equally, which may lead to noisy alignments due to poor initial representations. Inspired by curriculum learning \cite{wang2021survey}, a more effective approach is to start with easy samples and gradually incorporate harder ones. Instead of manually selecting confident samples via hard thresholds in the cost matrix, we reformulate the selection as a total mass constraint in Eq.~\eqref{eq:3}, eliminating the need for sensitive hyperparameter tuning. This results in the following formulation:
\begin{equation}
\label{eq:4}
\begin{aligned}
    \min_{Q \in \Pi} \quad & \langle Q, C \rangle_F + \lambda\, \mathrm{KL}\left(Q^\top \mathbf{1}_N \,\|\, \tfrac{\rho}{K} \mathbf{1}_K\right) \\
    \text{subject to} \quad & \Pi = \left\{ Q \in \mathbb{R}_+^{N \times K} \;\middle|\; Q \mathbf{1}_K \leq \tfrac{1}{N} \mathbf{1}_N,\; \mathbf{1}_N^\top Q \mathbf{1}_K = \rho \right\}
\end{aligned}
\end{equation}
where \( \rho \in (0, 1] \) is the fraction of selected mass, which gradually increases during training. The resulting \( Q \mathbf{1}_K \) represents the transport weights for each instance. We employ a sigmoid ramp-up function, a common technique in semi-supervised learning \cite{samuli2017temporal,tarvainen2017mean}, to progressively increase \( \rho \) during training:
\begin{equation}
\label{eq:5}
    \rho = \rho_0 + \left(1 - \rho_0\right) \cdot e^{-5\left(1 - t/ \left(T\cdot I \right)\right)^2},
\end{equation}
where \( t \) is the current iteration, \( T \) is the ramp-up epochs and  \( I \)  is the number of iterations per epoch.

\myparagraph{Heterogeneity-aware OT.} 
We denote this OT formulation as heterogeneity-aware OT, as it incorporates dual marginal distribution constraints. By iteratively solving Eq.~\eqref{eq:4} with a sigmoid ramp-up of \( \rho \) during training,  this approach seamlessly integrates instance selection, weighting, and aggregation within a unified OT framework.

\subsection{Heterogeneity-aware OT Solver}

Existing scaling algorithms \cite{knight2008sinkhorn} are designed for unbalanced OT, whereas our heterogeneity-aware OT, with its two tailored marginal constraints, differs from standard unbalanced OT. Despite this, by incorporating a virtual point, we can recast our formulation into an unbalanced OT problem that these efficient algorithms can solve.

As shown in Fig.~\ref{fig_OTSurv}(b), we introduce a \textit{virtual survival token} to absorb the \(1-\rho\) unselected mass, ensuring compatibility with the unbalanced OT framework. Specifically, we augment the cost matrix \(C\) with an additional column of zeros to form \(\hat{C} \in \mathbb{R}_+^{N \times (K+1)}\). Consequently, Eq.~\eqref{eq:4} can be reformulated as follows:
\begin{equation}
\label{eq:6}
    \begin{aligned}
        \min_{\hat{Q} \in \Phi} \quad & \langle \hat{Q}, \hat{C} \rangle_F 
        + \hat{\text{KL}}\left(\hat{Q}^\top \mathbf{1}_N, \beta, \hat{\lambda}\right) \\
        \text{subject to} \quad & \Phi = \left\{ \hat{Q} \in \mathbb{R}_+^{N \times \left(K+1\right)} 
        \mid \hat{Q} \mathbf{1}_{K+1} = \frac{1}{N} \mathbf{1}_N \right\}
    \end{aligned}
\end{equation}
where 
\begin{equation}
\label{eq:7}
    \begin{aligned}
        & \hat{C} = [C, \mathbf{0}_N], \quad \beta = \begin{bmatrix} \frac{\rho}{K} \mathbf{1}_K \\ 1 - \rho \end{bmatrix}, \quad \hat{\lambda} = \begin{bmatrix} \lambda \mathbf{1}_K \\ +\infty \end{bmatrix}, \\
        & \hat{\text{KL}}\left(\hat{Q}^\top \mathbf{1}_N, \beta, \hat{\lambda} \right) = 
        \sum_{i=1}^{K+1} \hat{\lambda}_i \left[ \hat{Q}^\top \mathbf{1}_N \right]_i 
        \log \frac{[\hat{Q}^\top \mathbf{1}_N]_i}{\beta_i}.
    \end{aligned}
\end{equation}
Here, the weighted KL divergence enforces that the virtual survival token absorbs exactly \(1-\rho\) of the unselected mass. \cite{zhang2024p} theoretically shows that the optimal transport plan \( Q^\star \) from Eq.~\eqref{eq:4} corresponds to the first $K$ columns of the optimal transport plan \( \hat{Q}^\star \) from Eq.~\eqref{eq:6}. The pseudocode is provided in Alg.~\ref{alg:srot}, where \(\oslash\) denotes element-wise division and \( \circ \) denotes Hadamard power, \ie, element-wise exponentiation. Notably, all operations in Alg.~\ref{alg:srot} are differentiable, making Heterogeneity-aware OT suitable for end-to-end training.

\begin{algorithm}[t!]
\caption{Scaling Algorithm for heterogeneity-aware OT}\label{alg:srot}
\SetAlgoLined
\footnotesize
\DontPrintSemicolon
\KwIn{Cost matrix $C$, $\epsilon$, $\lambda$, $\rho$, $N,K$, a large value $\iota$}    
$C \leftarrow [C, \mathbf{0}_N],\ 
\lambda \leftarrow [\lambda, ..., \lambda, \iota]^\top,\ 
\beta \leftarrow [\frac{\rho}{K} \mathbf{1}_K^\top, 1 - \rho]^\top$\;
$\alpha \leftarrow \frac{1}{N}\mathbf{1}_N,\ 
b \leftarrow 1_{K+1},\ 
M \leftarrow \exp(-C/\epsilon),\ 
f \leftarrow \lambda \oslash (\lambda + \epsilon)$\;
\While{$b$ not converge}{$a \leftarrow \alpha \oslash (Mb);\ 
b \leftarrow (\beta \oslash (M^\top a))^{\circ f}$\;}
$Q \leftarrow \text{diag}(a) M \text{diag}(b)$\;
\KwRet $Q[:, :K]$\;
\end{algorithm}

\section{Experiments}

\subsection{Implementation Details}

\myparagraph{Datasets.} 
We evaluate our OTSurv on six public TCGA datasets: BLCA (n=359), BRCA (n=868), LUAD (n=412), STAD (n=318), CRC (n=296), and KIRC (n=340), following the data split in \cite{songmultimodal}. WSIs are cropped into 256×256 non-overlapping patches at 20× magnification (averaging 12,789 patches per slide). Patch features extracted by UNI \cite{chen2024towards} (1024-D) are projected to 256-D via the projector. Survival prediction uses disease-specific survival \cite{liu2018integrated} as the label and is evaluated via 5-fold site-stratified cross-validation \cite{howard2021impact} with the concordance index (C-index).

\myparagraph{Setup.} 
We initialize the transport mass ratio at \(\rho_0 = 0.1\) and progressively increase it to 1.0 in 10 epochs using a sigmoid schedule. The number of survival tokens is set to \(K=16 \). Entropy regularization is set to \(\lambda = 0.1\). OTSurv is trained for 50 epochs using AdamW, with an initial learning rate of $1\times10^{-4}$ following cosine decay and a weight decay of \(1\times10^{-5}\). The Cox loss is optimized with a batch size of 16. All experiments are on an RTX 3090 GPU (24G).

\begin{table}[th]
\centering
\caption{Comparison of C-index Performance Across 6 TCGA Datasets for Different Survival Prediction Methods.}
\label{tab:combined}
\renewcommand{\arraystretch}{1.0} 
\begin{adjustbox}{width=\textwidth,center}
\begin{tabular}{@{}c@{\hskip 10pt}c@{\hskip 10pt}c@{\hskip 10pt}c@{\hskip 10pt}c@{\hskip 10pt}c@{\hskip 10pt}c@{\hskip 10pt}c@{}}
\toprule
\diagbox{\textbf{Method}}{\textbf{Dataset}} & \textbf{BRCA}          & \textbf{BLCA}          & \textbf{LUAD}          & \textbf{STAD}          & \textbf{CRC}           & \textbf{KIRC}          & \textbf{Mean} \\ 
\midrule
\textbf{Mean-Pooling}      & $0.512$\scriptsize$\pm 0.177$ & $0.589$\scriptsize$\pm 0.046$ & $0.532$\scriptsize$\pm 0.060$ & $0.503$\scriptsize$\pm 0.060$ & $0.597$\scriptsize$\pm 0.146$ & $0.730$\scriptsize$\pm 0.062$ & $0.577$ \\
\textbf{CLAM \cite{lu2021data}} & $0.629$\scriptsize$\pm 0.180$ & $0.608$\scriptsize$\pm 0.104$ & $0.569$\scriptsize$\pm 0.045$ & $0.541$\scriptsize$\pm 0.077$ & $0.625$\scriptsize$\pm 0.132$ & $0.674$\scriptsize$\pm 0.124$ & $0.608$ \\
\textbf{HIPT \cite{9880275}} & $0.555$\scriptsize$\pm 0.094$ & \textbf{$0.609$\scriptsize$\pm 0.127$} & $0.576$\scriptsize$\pm 0.081$ & $0.539$\scriptsize$\pm 0.074$ & $0.599$\scriptsize$\pm 0.066$ & $0.689$\scriptsize$\pm 0.094$ & $0.595$ \\
\textbf{ABMIL \cite{ilse2018attention,li2023task}} & $0.567$\scriptsize$\pm 0.092$ & $0.551$\scriptsize$\pm 0.055$ & $0.564$\scriptsize$\pm 0.044$ & \textbf{$0.561$\scriptsize$\pm 0.043$} & $0.657$\scriptsize$\pm 0.117$ & $0.675$\scriptsize$\pm 0.121$ & $0.596$ \\
\textbf{TransMIL \cite{shao2021transmil}} & $0.599$\scriptsize$\pm 0.058$ & $0.590$\scriptsize$\pm 0.106$ & $0.546$\scriptsize$\pm 0.117$ & $0.500$\scriptsize$\pm 0.061$ & $0.543$\scriptsize$\pm 0.127$ & $0.677$\scriptsize$\pm 0.133$ & $0.576$ \\
\textbf{AttnMISL \cite{yao2020whole}} & $0.585$\scriptsize$\pm 0.073$ & $0.523$\scriptsize$\pm 0.080$ & \textbf{$0.624$\scriptsize$\pm 0.139$} & $0.544$\scriptsize$\pm 0.056$ & \textbf{$\textbf{0.725}$\scriptsize$\pm 0.110$} & $0.658$\scriptsize$\pm 0.115$ & $0.610$ \\
\textbf{IB-MIL \cite{lin2023interventional}} & $0.511$\scriptsize$\pm 0.085$ & $0.536$\scriptsize$\pm 0.075$ & $0.594$\scriptsize$\pm 0.130$ & $0.541$\scriptsize$\pm 0.079$ & $0.576$\scriptsize$\pm 0.072$ & $0.666$\scriptsize$\pm 0.130$ & $0.571$ \\
\textbf{ILRA \cite{xiang2023exploring}} & $0.611$\scriptsize$\pm 0.135$ & $0.569$\scriptsize$\pm 0.082$ & $0.515$\scriptsize$\pm 0.063$ & $0.554$\scriptsize$\pm 0.060$ & $0.648$\scriptsize$\pm 0.123$ & $0.649$\scriptsize$\pm 0.101$ & $0.591$ \\
\textbf{PANTHER \cite{song2024morphological}} & \textbf{$\textbf{0.650}$\scriptsize$\pm 0.139$} & $0.590$\scriptsize$\pm 0.084$ & $0.575$\scriptsize$\pm 0.046$ & $0.504$\scriptsize$\pm 0.082$ & $0.641$\scriptsize$\pm 0.133$ & \textbf{$0.702$\scriptsize$\pm 0.124$} & $0.610$ \\
\midrule
\textbf{OTSurv}    & $0.621$\scriptsize$\pm 0.071$ & $\textbf{0.637}$\scriptsize$\pm 0.065$ & $\textbf{0.638}$\scriptsize$\pm 0.077$ & $\textbf{0.565}$\scriptsize$\pm 0.057$ & $0.667$\scriptsize$\pm 0.111$ & $\textbf{0.750}$\scriptsize$\pm 0.149$ & $\textbf{0.646}$ \\
\bottomrule
\end{tabular}
\end{adjustbox}
\vspace{-10pt}
\end{table}

\subsection{Comparison with State-of-the-Art Methods}
In Table~\ref{tab:combined}, across six TCGA cancer datasets, OTSurv consistently outperforms Mean-Pooling, CLAM \cite{lu2021data}, HIPT \cite{9880275}, ABMIL \cite{ilse2018attention,li2023task}, TransMIL \cite{shao2021transmil}, AttnMISL \cite{yao2020whole}, IB-MIL \cite{lin2023interventional}, ILRA \cite{xiang2023exploring}, and PANTHER \cite{song2024morphological}. Moreover, we perform log-rank tests \cite{bland2004logrank} on high- and low-risk cohorts, defined by splitting the predicted risks at the median (50\%). Fig.~\ref{fig_OTSurv_heatmap}(a) presents Kaplan-Meier survival curves along with p-values for six cancer types, all of which are below the 0.05 significance threshold, thereby demonstrating effective risk stratification.

\subsection{Ablation Studies}

\myparagraph{Design choices.}
Our ablation study, as in Table~\ref{tab:ablation_combined}, shows that replacing the KL-based global constraint with an equality constraint (\textbf{$\mathbf{C_G}$ \ding{55}}) and replacing the progressive partial local constraint with either a fixed partial constraint ($\mathbf{C_L}$ \ding{55} with fixed $\rho=0.8$) or a fixed full constraint ($\mathbf{C_L}$ \ding{55} with fixed $\rho=1.0$) both degrade performance -- highlighting the necessity of both \(\mathbf{C_G}\) and \(\mathbf{C_L}\). We also evaluated several other model design choices:
{(i)} Replacing our mass-based instance selection with a cost matrix-based hard thresholding method leads to performance degradation.
{(ii)} Replacing the sigmoid ramp-up of $\rho$ with a linear ramp-up remains effective;
{(iii)} Substituting Cox loss with NLL survival loss \cite{zadeh2020bias} yields inferior performance, likely due to information loss when converting regression to classification;
{(iv)} Omitting the linear projector and training on raw patch embeddings significantly reduces performance, underscoring the importance of dimension reduction.

\myparagraph{OT Hyperparameters.}  
We evaluate OT hyperparameters -- including the initial mass ratio \( \rho_0 \), KL weight \( \lambda \), ramp-up epochs \( T \), the number of survival tokens \( K \) and the number of sampled patches per WSI  \( N \). Our results show that moderate variations have little impact, whereas extreme settings degrade performance, supporting our design choices.

\begin{table}[t]
\centering
\caption{Ablations on model design choices and OT hyperparameters.}
    
\label{tab:ablation_combined}
\renewcommand{\arraystretch}{0.90} 
\begin{adjustbox}{width=\textwidth,center}

\begin{tabular}{@{}c@{\hskip 5pt}c@{\hskip 10pt}c@{\hskip 10pt}c@{\hskip 10pt}c@{\hskip 10pt}c@{\hskip 10pt}c@{\hskip 10pt}c@{\hskip 10pt}c@{}}
\toprule

\multicolumn{2}{c}{\multirow{2}{*}{\textbf{Method}} }
& \multicolumn{6}{c}{\textbf{Dataset}} 
& \multirow{2}{*}{\textbf{Mean}} \\
\cmidrule(lr){3-8}
 &  & \textbf{BRCA} & \textbf{BLCA} & \textbf{LUAD} 
  & \textbf{STAD} & \textbf{CRC} & \textbf{KIRC} &  \\

\midrule

\multicolumn{2}{c}{\textbf{OTSurv}}
& $0.621$\scriptsize$\pm 0.071$ & $0.637$\scriptsize$\pm 0.065$ & $0.638$\scriptsize$\pm 0.077$ & $0.565$\scriptsize$\pm 0.057$ & $0.667$\scriptsize$\pm 0.111$ & $0.750$\scriptsize$\pm 0.149$ & $\textbf{0.646}$ \\     

\midrule
\multicolumn{8}{c}{\textbf{(1) Ablation on model  design choices}} \\
\midrule

\multirow{3}{*}{\makecell{\makecell{\textbf{$\mathbf{C_G}$ \ding{51}}}}}
& $\mathbf{C_L}$ \ding{51} (Sigmoid $\rho$)     
    & $0.621$\scriptsize$\pm 0.071$ & $0.637$\scriptsize$\pm 0.065$ & $0.638$\scriptsize$\pm 0.077$ & $0.565$\scriptsize$\pm 0.057$ & $0.667$\scriptsize$\pm 0.111$ & $0.750$\scriptsize$\pm 0.149$ & $ 0.646$ \\ 
& $\mathbf{C_L}$  \ding{55} (Fix $\rho=0.8$)   
    & $0.646$\scriptsize$\pm 0.106$ & $0.628$\scriptsize$\pm 0.065$ & $0.631$\scriptsize$\pm 0.080$ & $0.547$\scriptsize$\pm 0.051$ & $0.631$\scriptsize$\pm 0.183$ & $0.747$\scriptsize$\pm 0.134$ & $0.638$ \\ 
& $\mathbf{C_L}$ \ding{55}  (Fix $\rho=1.0$)   
    & $0.556$\scriptsize$\pm 0.185$ & $0.628$\scriptsize$\pm 0.064$ & $0.637$\scriptsize$\pm 0.071$ & $0.551$\scriptsize$\pm 0.046$ & $0.616$\scriptsize$\pm 0.127$ & $0.746$\scriptsize$\pm 0.135$ & $0.622$ \\  
\midrule

\multirow{3}{*}{\makecell{\makecell{\textbf{$\mathbf{C_G}$ \ding{55}}}}}
& $\mathbf{C_L}$ \ding{51} (Sigmoid $\rho$)                            & $0.617$\scriptsize$\pm 0.131$ & $0.604$\scriptsize$\pm 0.104$ & $0.637$\scriptsize$\pm 0.097$ & $0.553$\scriptsize$\pm 0.064$ & $0.611$\scriptsize$\pm 0.148$ & $0.724$\scriptsize$\pm 0.124$ & $0.624$ \\ 
& $\mathbf{C_L}$  \ding{55} (Fix $\rho=0.8$)             & $0.566$\scriptsize$\pm 0.199$ & $0.592$\scriptsize$\pm 0.104$ & $0.638$\scriptsize$\pm 0.062$ & $0.551$\scriptsize$\pm 0.033$ & $0.624$\scriptsize$\pm 0.148$ & $0.737$\scriptsize$\pm 0.179$ & $0.618$ \\ 
& $\mathbf{C_L}$ \ding{55}  (Fix $\rho=1.0$)              & $0.635$\scriptsize$\pm 0.102$ & $0.590$\scriptsize$\pm 0.119$ & $0.646$\scriptsize$\pm 0.074$ & $0.501$\scriptsize$\pm 0.119$ & $0.530$\scriptsize$\pm 0.160$ & $0.745$\scriptsize$\pm 0.145$ & $0.608$  \\
\midrule

\multirow{1}{*}{\makecell{\textbf{Patch Select.}}}   
& Mass $\to$  Cost                    & $0.589$\scriptsize$\pm 0.136$ & $0.612$\scriptsize$\pm 0.079$ & $0.637$\scriptsize$\pm 0.048$ & $0.515$\scriptsize$\pm 0.093$ & $0.559$\scriptsize$\pm 0.148$ & $0.722$\scriptsize$\pm 0.171$ & $0.606$ \\
\midrule

\multirow{1}{*}{\makecell{\textbf{Ramp-up}}}   
& Sig. $\to$  Lin.                    & $0.675$\scriptsize$\pm 0.113$ & $0.632$\scriptsize$\pm 0.055$ & $0.639$\scriptsize$\pm 0.072$ & $0.547$\scriptsize$\pm 0.046$ & $0.632$\scriptsize$\pm 0.117$ & $0.742$\scriptsize$\pm 0.134$ & $0.645$ \\    
\midrule

\multirow{1}{*}{\makecell{\textbf{Loss}}} 
& Cox $\to$  NLL                                & $0.624$\scriptsize$\pm 0.117$ & $0.609$\scriptsize$\pm 0.112$ & $0.651$\scriptsize$\pm 0.084$ & $0.565$\scriptsize$\pm 0.040$ & $0.606$\scriptsize$\pm 0.163$ & $0.762$\scriptsize$\pm 0.125$ & $0.636$ \\   
\midrule

\multirow{1}{*}{\makecell{\textbf{Projector}}} 
& w/o Proj.                                 & $0.601$\scriptsize$\pm 0.127$ & $0.587$\scriptsize$\pm 0.062$ & $0.604$\scriptsize$\pm 0.059$ & $0.518$\scriptsize$\pm 0.098$ & $0.544$\scriptsize$\pm 0.097$ & $0.724$\scriptsize$\pm 0.145$ & $0.596$ \\
\midrule

\multicolumn{8}{c}{\textbf{(2) OT Hyperparameters}} \\
\midrule

\multirow{2}{*}{\makecell{ \textbf{Initial Mass} \\ \textbf{Ratio $\rho_0$}}} 
& 0.1 $\to$  0.2                               & $0.638$\scriptsize$\pm 0.118$ & $0.579$\scriptsize$\pm 0.092$ & $0.656$\scriptsize$\pm 0.109$ & $0.579$\scriptsize$\pm 0.054$ & $0.606$\scriptsize$\pm 0.113$ & $0.750$\scriptsize$\pm 0.156$ & $0.635$ \\     
& 0.1 $\to$  0.05                                & $0.639$\scriptsize$\pm 0.119$ & $0.570$\scriptsize$\pm 0.077$ & $0.666$\scriptsize$\pm 0.111$ & $0.561$\scriptsize$\pm 0.077$ & $0.570$\scriptsize$\pm 0.094$ & $0.761$\scriptsize$\pm 0.147$ & $0.628$ \\
\midrule

\multirow{1}{*}{\makecell{ \textbf{KL Weight $\lambda$}}} 
& 0.1 $\to$ 0.2                                & $0.615$\scriptsize$\pm 0.159$ & $0.621$\scriptsize$\pm 0.056$ & $0.583$\scriptsize$\pm 0.048$ & $0.552$\scriptsize$\pm 0.102$ & $0.685$\scriptsize$\pm 0.192$ & $0.755$\scriptsize$\pm 0.113$ & $0.635$ \\
\midrule

\multirow{2}{*}{\makecell{ \textbf{Ramp-up} \\ \textbf{Epoches $T$}}} 
& 10 $\to$ 20                                & $0.626$\scriptsize$\pm 0.105$ & $0.612$\scriptsize$\pm 0.112$ & $0.645$\scriptsize$\pm 0.081$ & $0.560$\scriptsize$\pm 0.040$ & $0.606$\scriptsize$\pm 0.159$ & $0.773$\scriptsize$\pm 0.125$ & $0.637$ \\
& 10 $\to$ 30                                & $0.662$\scriptsize$\pm 0.122$ & $0.640$\scriptsize$\pm 0.032$ & $0.556$\scriptsize$\pm 0.062$ & $0.527$\scriptsize$\pm 0.048$ & $0.612$\scriptsize$\pm 0.085$ & $0.776$\scriptsize$\pm 0.120$ & $0.629$ \\ 
\midrule

\multirow{2}{*}{\makecell{ \textbf{Survival} \\ \textbf{Token $K$}}} 
& 16 $\to$ 8                              & $0.659$\scriptsize$\pm 0.085$ & $0.630$\scriptsize$\pm 0.066$ & $0.636$\scriptsize$\pm 0.068$ & $0.548$\scriptsize$\pm 0.053$ & $0.619$\scriptsize$\pm 0.121$ & $0.767$\scriptsize$\pm 0.115$ & $0.643$ \\   
& 16 $\to$ 32                             & $0.644$\scriptsize$\pm 0.087$ & $0.628$\scriptsize$\pm 0.069$ & $0.636$\scriptsize$\pm 0.071$ & $0.570$\scriptsize$\pm 0.053$ & $0.621$\scriptsize$\pm 0.123$ & $0.761$\scriptsize$\pm 0.136$ & $0.643$ \\  

\midrule
\multirow{1}{*}{\makecell{ \textbf{Patch $N$}}} 
& All $\to$ 3000                              & $0.646$\scriptsize$\pm 0.106$ & $0.628$\scriptsize$\pm 0.065$ & $0.631$\scriptsize$\pm 0.080$ & $0.547$\scriptsize$\pm 0.051$ & $0.631$\scriptsize$\pm 0.183$ & $0.747$\scriptsize$\pm 0.134$ & $0.638$ \\

\bottomrule
\end{tabular}
\end{adjustbox}
\vspace{-10pt}
\end{table}

\subsection{Model Interpretation}

\begin{figure*}[!t]
\begin{center}
\includegraphics[width=0.90\linewidth]{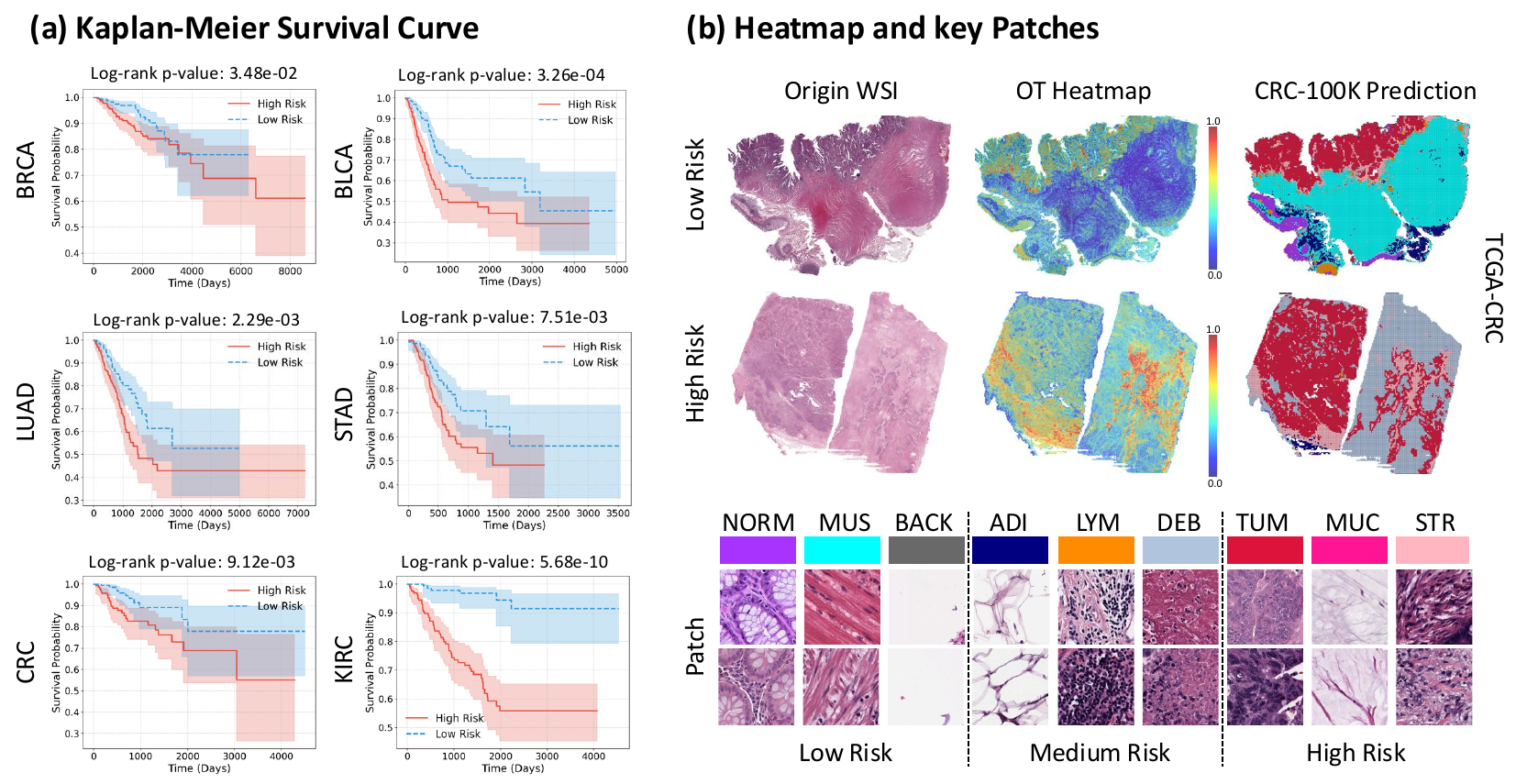}
\end{center}
\vspace{-18pt}
\caption{(a) Kaplan–Meier survival curves, (b) OTSurv interpretability analysis.}
\label{fig_OTSurv_heatmap}
\vspace{-10pt}
\end{figure*}

In Fig.~\ref{fig_OTSurv_heatmap}(b), we compare the heatmap of OTSurv with tissue-type map. The heatmap is computed as \( \text{Attention} = Q \cdot \left| W_{\text{agg}}^\top  \right| \), where \(Q \in \mathbb{R}^{N \times K}\) is the transport matrix and \(W_{\text{agg}} \in \mathbb{R}^{1 \times K}\) represents the aggregation weight of each survival token in $f_{\text{agg}}$. The tissue-type map is generated using a linear classifier trained on the CRC-100K dataset (F1 = 0.93, AUC = 0.89) with patch features extracted by UNI \cite{chen2024towards}. Besides, we average patch-level attentions across 9 tissue types in all patches of the TCGA-CRC dataset. As shown in the bottom row of Fig.~\ref{fig_OTSurv_heatmap}(b), patches positioned further to the right have higher attention values (\ie, greater impact on survival prediction), aligning with clinical findings. This demonstrates that OTSurv has the capability to capture biologically relevant patterns and support interpretable risk stratification.

\section{Conclusion}

We introduced OTSurv, a novel optimal transport-based MIL method for survival prediction that explicitly models dual-scale pathological heterogeneity -- mirroring how pathologists assess diverse tissue patterns. Our approach formulates survival prediction as a heterogeneity-aware OT problem by incorporating a global long-tail constraint to model morphological distributions and a local uncertainty-aware constraint to select high-confidence patches. We then reformulate the problem as an unbalanced OT task, solved efficiently via a matrix scaling algorithm. Experiments show that OTSurv achieves state-of-the-art performance and captures biologically meaningful features.

%
%
\bibliographystyle{splncs04.bst}



\end{document}